\pdfoutput=1
\documentclass[11pt]{article}
\usepackage{acl}
\usepackage{times}
\usepackage{latexsym}
\usepackage[T1]{fontenc}
\usepackage[utf8]{inputenc}
\usepackage{microtype}

\usepackage{float}

\usepackage{graphicx, float}
\usepackage{todonotes}
\usepackage{subfig}
\usepackage{times}
\usepackage{latexsym}
\usepackage[T1]{fontenc}
\usepackage[utf8]{inputenc}
\usepackage{microtype}

\usepackage{hyperref}

\usepackage{amsmath}
\usepackage{booktabs}
\usepackage[a]{esvect}

\title{Cross-lingual Similarity of Multilingual Representations Revisited}

\author{Maksym Del \and Mark Fishel \\
  Institute of Computer Science \\
  University of Tartu, Estonia \\
  \texttt{\{maksym,mark\}@tartunlp.ai} 
  }

\begin{document}
\maketitle

\begin{abstract}
% The similarity of neural network representations is an ill-defined concept, and different similarity indexes might be better or worse depending on the goal.
% Our work addresses a similarity of neural network representations and metrics used to measure it.
Related works used indexes like CKA and variants of CCA to measure the similarity of cross-lingual representations in multilingual language models. In this paper, we argue that assumptions of CKA/CCA align poorly with one of the motivating goals of cross-lingual learning analysis, i.e., explaining zero-shot cross-lingual transfer. We highlight what valuable aspects of cross-lingual similarity these indexes fail to capture and provide a motivating case study \textit{demonstrating the problem empirically}. Then, we introduce \textit{Average Neuron-Wise Correlation (ANC)} as a straightforward alternative that is exempt from the difficulties of CKA/CCA and is good specifically in a cross-lingual context.
Finally, we use ANC to construct evidence that the previously introduced ``first align, then predict'' pattern takes place not only in masked language models (MLMs) but also in multilingual models with \textit{causal language modeling} objectives (CLMs). Moreover, we show that the pattern extends to the \textit{scaled versions} of the MLMs and CLMs (up to 85x original mBERT).\footnote{Our code is publicly available at \url{https://github.com/TartuNLP/xsim}}

\end{abstract}

\section{Introduction}

Similarity indexes like Canonical Correlation Analysis (CCA, \citealp{cca-Ramsay1984MatrixC}) or Centered Kernel Alignment (CKA, \citealp{cka=pmlr-v97-kornblith19a}) aim to find a similarity between parallel sets of different representations of the same data. The deep learning community adapted these indexes to measure similarity between representations that \textit{come from different models} \cite{svcca-NIPS2017_7188, pwcca-NIPS2018_7815, cka=pmlr-v97-kornblith19a}. 
Another line of work used the same methods to measure similarity \textit{between different languages} which come from a \textit{single} multilingual model \cite{mt-svcca, Singh2019BERTIN, conneau-etal-2020-emerging, muller-etal-2021-first}. 

In this paper, we argue that while CCA/CKA methods are a good fit for the first case, they are a suboptimal choice for the second scenario.

First, we employ a real-world motivating example to demonstrate that CKA can fail to capture the notion of similarity that we consider helpful in a cross-lingual context. We also discuss the general problems of CKA/CCA indexes and conclude that they are not well aligned with some of the goals of cross-lingual analysis \textbf{(Section~\ref{sec:motiv})}.

Next, we propose and verify an Averaged Neuron-Wise Correlation (ANC) as a straightforward alternative. % that avoids the issues with CKA/CCA.
It exploits the fact that representations from the same model have apriori-aligned neurons, which is the desired property in a cross-lingual setup \textbf{(Section~\ref{sec:method})}.

Finally, \citet{muller-etal-2021-first} demonstrated the so-called ``first align, then predict'' representational pattern in a multilingual model: the model first aligns representations of different languages together, and then (starting from the middle layers) makes them more language-specific again (to accompany the language-specific training objective). The finding is insightful but only considers mBERT \cite{wu-dredze-2019-beto} which is a masked language model (MLM) with 110M parameters. Thus, it is unclear if the ``first align, then predict'' pattern is specific to this model or more general. In this study, we use ANC to show that the pattern generalizes to the GPT-style \cite{gpt-NEURIPS2020_1457c0d6} causal language models (CLMs, \citealp{xglm}) and extends to   
\textit{large-scale} MLMs and CLMs \textbf{(Section \ref{sec:scaling_laws})}.

In this paper we are interested specifically in the scenario of measuring the strength of cross-lingual similarity of representations that come from a single multilingual language model. This scenario is very common in the field as it is often not feasable to train a separate models for each language and we present a method that allows for better representational similarity analysis then CKA/CCA.

\newpage
In summary, our contributions are three-fold:
\begin{itemize}
    \item conceptual and \textit{empirical} critique of CKA/CCA for cross-lingual similarity analysis \textbf{(Section~\ref{sec:motiv})};
    \item \textit{Average Neuron-Wise Correlation} as a simple alternative method designed specifically for cross-lingual similarity \textbf{(Section~\ref{sec:method})};
    \item \textit{scaling laws} of cross-lingual similarity in both multilingual MLMs and CLMs \textbf{(Section~\ref{sec:scaling_laws})}.
\end{itemize}

\section{Related work}
\label{sec:relwork}
\citet{cca-Ramsay1984MatrixC} introduced CCA as a method for measuring canonical correlations between two sets of random variables. \citet{svcca-NIPS2017_7188} proposed a variant of the CCA called SVCCA and used it to analyze representations \textit{between different neural networks}.  \citet{pwcca-NIPS2018_7815} proposed PWCCA, another improvement to CCA for the network analysis, and \citet{cka=pmlr-v97-kornblith19a} analyzed CCA, SVCCA, PWCCA, and other methods concluding that CKA is superior to them. 

In a cross-lingual setting, we have a single network, and we compare representations that come from different languages. Following the introduction of SVCCA, \citet{mt-svcca} used it to compare language representations (at different layers) in a multilingual neural machine translation system. The method we present in this work applies to the seq2seq models, but in this work, we focus on models trained with CLM and MLM objectives while leaving seq2seq for future work. \citet{Singh2019BERTIN} performed a similar study where they focused on the multilingual BERT model\footnote{\url{https://github.com/google-research/bert/blob/master/multilingual.md}}  and employed PWCCA as a similarity index. The conclusion was that language representations diverge with network depth. 

On the other hand, \citet{conneau-etal-2020-emerging} and \citet{muller-etal-2021-first} used CKA and behavior analysis to show that the opposite pattern takes place: language representations align with the network depth and only moderately decrease towards the end. In other words, representations first converge towards language neutrality and then recover some language-specificity. The alignment makes zero-shot cross-lingual transfer possible, and slight divergence accompanies language-specific training objectives (such as English downstream prediction task or predicting words in the particular language as in masked language modeling objective). Following \citealt{muller-etal-2021-first}, we call this phenomenon the \textit{``first align, then predict''} pattern. 

Eventually, \citet{del-baltic} showed that the similarity analysis was different because \citet{Singh2019BERTIN} used CLS-pooling while \citet{muller-etal-2021-first} used mean-pooling to convert token embeddings into a sentence representation. They also showed that mean-pooling is a better option.

Finally, \citet{Li2015ConvergentLD} aligned most correlated neurons between layers of two different networks and then computed similarity from the recovered correspondence. The method we propose in this paper is similar in spirit to this one, except we focus on the cross-lingual analysis of multilingual models and thus have no need to find the alignment between neurons.

In this work, we build on these studies in three ways: we demonstrate that even CKA can fail to provide relevant cross-lingual similarity, we propose another method to compare multilingual representations, and we reveal that the ``first align, then predict'' pattern generalizes across training objectives and holds for models of large sizes.

\section{Similarity Indexes Background}
\label{sec:bg}
In this section, we provide some background on CKA and CCA, SVCCA, and PWCCA similarity indexes\footnote{In the paper, we refer to both SVCCA and PWCCA simply as CCA unless otherwise specified.}. We focus on the parts of the methods most relevant to the key points we make in this work. For the full mathematical description refer to \citet{cka=pmlr-v97-kornblith19a}.

\paragraph{Neuron} Following related works, we define a neuron as a vector of values it takes over a dataset \cite{Li2015ConvergentLD, svcca-NIPS2017_7188, pwcca-NIPS2018_7815, cka=pmlr-v97-kornblith19a}. Formally, let $D$ be a dataset consisting of data examples $\vv{d}$: 
\[ D = \{\vv{d_1}, \cdots \vv{d_m} \} \] 
Let $\varphi_i$ be a function that returns a neuron activation value for the training example at the $i$-th unit of the $l$-th layer of the network. The \textit{neuron} $\vv{z}_i$ is the \textit{vector} of network activations recorded by applying $\varphi_i$ over the elements of $D$, i.e.
\[ \vv{z^i} = [\varphi_i(\vv{d_1}), \cdots , \varphi_i(\vv{d_m})] \]
In practice, we pass a set of data examples to the network and record activations for each unit at every layer. The vector of these activations is what we consider a representation of a neuron $\vv{z}$.

\paragraph{Layer} The frequent goal of representational similarity analysis is to compare layers of neural networks. Under our definition, the layer ${L}$ is the list of vectors (matrix) that consists of the \textit{neurons} at a particular depth, i.e. 
\[ {L} = [\vv{z^i}, \cdots , \vv{z^n}] \] where $n$ is the number of neurons at layer $L$.
Alternatively, we can think of layer $L$ as the subspace of $R^m$ spanned by its neurons $(\vv{z^i}, \cdots , \vv{z^n})$, where $m$ is the number of examples in the dataset.

CCA/CKA indexes rely on the idea of subspaces spanned by the neurons, making them powerful when comparing representations across \textit{different networks}. There can be more neurons in the first layer than in the second; the neurons also do not need to be aligned. CCA/CKA uses neurons only to describe the vector subspaces and then compare the subspaces as opposite to the neurons themselves. 

That is why methods like CKA and CCA  try to find some second-order descriptions of representational spaces (e.g., gram matrices/canonical vectors) and compare these. The decisions on what second-order information to consider and what comparison technique to use define the differences between the indexes.       

% This makes indexes possess various invariance properties that are desirable when comparing layers that come, e.g., from different networks. However, we discuss the implications of these properties for cross-lingual analysis in Section~\ref{sec:invariance}.  

\paragraph{Dominant Correlations}
\label{sec:bg-cka}
The first step for all methods is to center each neuron in the layer representations: 
\begin{center}
    $X:=L_1 - mean(L_1)$\\
    $Y:=L_2 - mean(L_2)$
\end{center}

Let X and Y have $p_1$ and $p_2$ neurons (columns). Consider gram matrix $XX^\text{T}$. Because neurons in X are centered,  $XX^\text{T}$ is proportional to covariance matrix of $X$. Therefore, the elements in $XX^\text{T}$ correspond to all pairwise covariance similarities data points in X (the same holds for $YY^\text{T}$).

Now consider doing eigendecomposition of $X^\text{T}X$. Eigenvectors $\vv{u}_X^i | i\in \{1, ..., m\}, \vv{u}_X^i\in R^m$ will represent directions of the most dominant correlations of data points in X. Also, we can think about vectors $\vv{u}_X^i$ as of \textit{eigenneurons}, the ones that explain the most variance in the representational space of other neurons. $\lambda_X^i$ is then the $i^\text{th}$ eigenvalue of $XX^\text{T}$ (the strengths of the eigenneurons).

\paragraph{CCA}

The directions $\vv{u}_X$ and $\vv{u}_Y$  are orthogonal by the definition of the eigendecomposition. The pair of vectors with the maximum dot product $\langle\vv{u}_X$, $\vv{u}_Y\rangle$ is called the first pair of canonical directions. The value of their dot product is the first CCA coefficient. Then the second pair produces the second canonical coefficient, and so on.

 The formula for the CCA similarity index is then as follows (from \citealp{cka=pmlr-v97-kornblith19a}):

\begin{align}
    CCA(XX^\text{T}, YY^\text{T}) = \sum_{i=1}^{p_1} \sum_{j=1}^{p_2} \langle \vv{u_X^i}, \vv{u_Y^j}\rangle^2 / p_1 .
    \label{eq:cca}
\end{align}

\paragraph{CKA} We might also consider weighting the CCA correlations by their eigenvalues. This results in Linear CKA (from \citealp{cka=pmlr-v97-kornblith19a}):
    $$\text{CKA}(XX^\text{T}, YY^\text{T}) =$$
\begin{align}
    &= \frac{\sum_{i=1}^{p_1} \sum_{j=1}^{p_2} \lambda_X^i \lambda_Y^j \langle \vv{u_X^i}, \vv{u_Y^j}\rangle^2}{\sqrt{\sum_{i=1}^{p_1} (\lambda_X^i)^2}\sqrt{\sum_{j=1}^{p_2} (\lambda_Y^j)^2}}
    \label{eq:cka}
\end{align}
In this work, we focus on Linear CKA because related works such as \citet{ muller-etal-2021-first} and \citet{conneau-etal-2020-emerging} use it.

\textbf{SVCCA} If we also decide to apply SVD as the preprocessing step after centering, we get SVCCA. CCA then computes correlation coefficients only for top K components from SVD transformed data (right singular values) and thus can be better averaged (see Equation \ref{eq:cca}).  

\textbf{PWCCA} Finally, instead of taking a simple average of CCA coefficients or weighting them by singular values (as in CKA), we might weight them weights (loosely speaking) related to the CCA directions that encapsulate the most data when projected.

In summary, all these methods are related and based on the idea that we can deduce some dominant correlation directions in $X$ and $Y$ and then compare these. Another way to look at it is that if CCA/CKA can represent neurons in $Y$ as linear combinations of neurons in $X$, these correlation methods will generally respond with high scores. 

The differences between methods make them invariant to the data scaling, centering, and orthogonal transformations. At the same time, CCA and SVCCA will not change their scores under any invertible linear transformations of either $X$ or $Y$ (see \citealp{cka=pmlr-v97-kornblith19a} for more details).

\section{Problems With CKA/CCA}
\label{sec:motiv}

By performing an illustrative experiment, let us introduce problems with CKA  and CCA indexes.

Specifically, we want to check if different normalization choices of the Transformer \cite{trans-NIPS2017_3f5ee243} layers impact the zero-shot cross-lingual transfer capabilities of the model and the similarity of cross-lingual representations it learns.

This section presents a two-fold case against CKA/CCA for cross-lingual similarity analysis:
\begin{itemize}
    \item empirical: CKA fails to uncover relationships between similarity after the architectural change that does not hurt the performance of the model;
    \item conceptual: lack of interpretability and unsatisfying underlying assumptions in CCA/CKA.
\end{itemize}

\subsection{Experiments Setup}  
\label{sec:motiv:setup}

\paragraph{Models} We train the following three XLM-Roberta \cite{xlmr} language models (\verb|base| size versions) from scratch  (each with a different normalization schema):

\begin{itemize}
    \item Post-LN (\verb|scale_post|): normalization block is placed \textit{after} the residual connections in the transformer block (part of the original Transfomer);
    \item Pre-LN (\verb|scale_pre|): normalization block is placed \textit{before} the residuals (this was shown to improve training by \citealp{preln-10.5555/3524938.3525913});
    \item Normformer (\verb|scale_normformer|): normalization block is placed \textit{before} the residuals \textit{and} FeedForward, Residual, and Self-Attention layers are also normalized \cite{normformer-https://doi.org/10.48550/arxiv.2110.09456}.
\end{itemize}

\paragraph{Pre-Training}
We pre-train a model based on XLM-R Base using 50M sentences uniformly sampled from four languages: English, French, Estonian, and Bulgarian. We chose the languages to be reasonably diverse: French is the most similar to English in both grammar and alphabet, Bulgarian is from a different language group (Slavic), and Estonian is from a completely different language family (Finno-Ugric). We train the model for 1M batches of 512 sentences from the \textit{CC100} dataset using two Nvidia A100 GPUs. The only architectural difference from the original XLM-Roberta is that we change normalization types to Pre-LN and Normformer; other setup details are painstakingly identical.

\paragraph{Experiment 1: XNLI Fine-Tuning}
After having three models pretrained, we fine-tune each of them on XNLI sentence classification task \cite{conneau2018xnli}. We use only English data for training but evaluate on English and other language evaluation sets (we only skip Estonian since it is not a part of XNLI). This setup, where we tune on one language but use another at test time, is called \textit{zero-shot cross-lingual transfer}.   

\paragraph{Experiment 2: CKA Similarirty}
\label{sec:setup-sim}
After having the XNLI zero-shot cross-lingual transfer scores, we extract sentence representations from all layers of each model and compare layers using the CKA similarity index.

 The parallel corpus is composed of \citet{singh-etal-2019-bert}'s extension of the XNLI dataset (10k examples for each pair)\footnote{Using XNLI for both fine-tuning and CKA analysis allows us to avoid domain mismatch scenarios entirely}.

We embed the source and target sentences with the models and perform mean-pooling over tokens at each layer for each language pair (as suggested by \citealp{del-baltic}). Next, we compare two parallel sets of sentence representations using the CKA similarity index to get a similarity score for each layer.

\paragraph{Experiment 3: Per-Layer Matching Accuracy}
Lastly, to get insight into some cross-lingual behavioral capabilities of representations at each layer, we analyze them with a sentence-matching probing task.

We use the same data and pooling strategy as in the CKA analysis. For each English sentence, we find the closest target sentence in the opposite language (out of all 10k targets) by cosine similarity. If this sentence is the actual parallel counterpart (translation) of the English sentence, we say the model got this English example correct. Then we compute the accuracy of this sentence matching as the ratio between correctly labeled English examples and the total number (10k) of English examples.

Throughout this work, we conduct experiments across languages sampled from the four language families: Germanic, Romance, Slavic, Baltic, and Finno-Ugric. While the results hold across the complete set of languages from our work, we showcase different subsets of languages from language families in different experiments to introduce more diversity while keeping the plots concise. 

\subsection{Experiments Results}
\label{sec:motiv:results}
\paragraph{Experiment 1: XNLI Fine-Tuning}
See Table~\ref{tab:xnli} for our models' zero-shot cross-lingual transfer performance on the XNLI validation set.

\begin{table}[H]
\small
\centering
\begin{tabular}{l}
\hline
\textbf{Normalization} \\
\hline
\verb|scale_post|  \\
\verb|scale_pre|  \\
\verb|scale_normformer|  \\ 
\hline
\end{tabular}
\begin{tabular}{ccc}
\hline
\textbf{en} & \textbf{fr} & \textbf{bg}\\
\hline
0.79 & 0.72 & 0.70 \\ 
0.81 & 0.72 & 0.72 \\ 
0.79 & 0.72 & 0.71 \\ 
\hline
\end{tabular}
\caption{Accuracy of XLM-Roberta Base Transformers pre-trained with different normalization schemes and fine-tuned on the English portion of the XNLI sentence classification task. The models show similar zero-shot cross-lingual transfer performance.}
\label{tab:xnli}
\end{table}

The Table shows that all three models achieve solid zero-shot transfer performance with a cross-lingual transfer gap of 7-9\%. We see no significant gains from the \verb|scale_pre| or \verb|scale_normformer|, but crucially we see no significant losses either.

\begin{figure}[]
    \centering
    \includegraphics[width=0.34\textwidth]{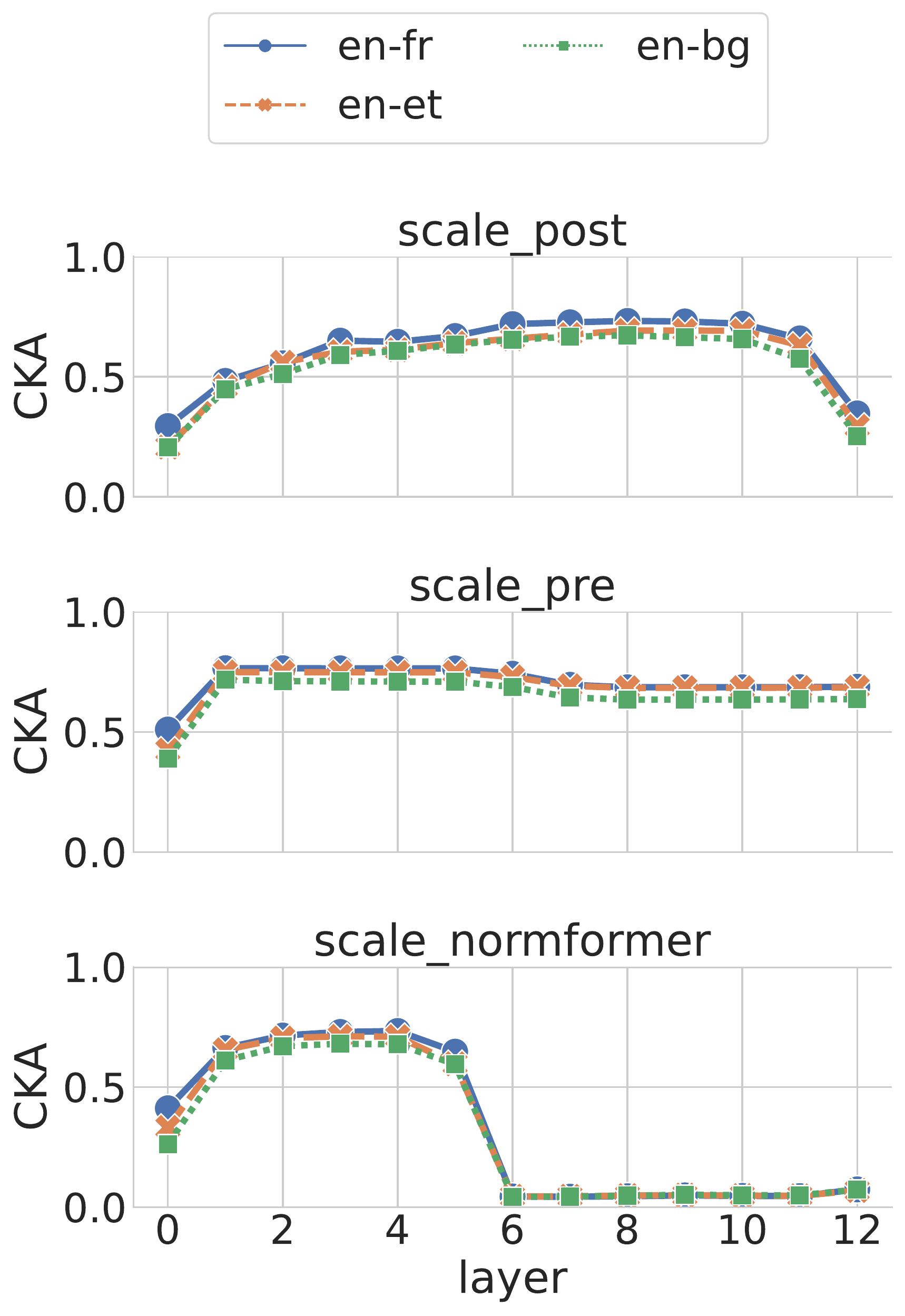}
    \caption{Motivating example 1: counter-intuitive CKA (dis)similarity of XLM-Normformer layers. CKA index shows drastic dissimilarity for layers 6-12 despite remarkable zero-shot cross-lingual transfer performance of the model.}
    \label{fig:normformer_cka}
\end{figure}

\paragraph{Experiment 2: CKA Similarirty}
We present per-layer CKA similarity results for the pre-trained (untuned) models in Figure \ref{fig:normformer_cka}.     

Figure \ref{fig:normformer_cka} reveals that while for \verb|scale_post| and \verb|scale_pre| CKA show fairly high cross-lingual performance at all layers, the Normformer results are drastically different. While the similarity for the first half of the layers increases (layers 0-5), the CKA score drops dramatically at the middle layer of the network and continues to hang around zero for all remaining layers (layers 6-12).

This result is especially surprising because CKA confidently gives similarity scores that are almost zero, while Table \ref{tab:xnli} shows no substantial difference in the zero-shot cross-lingual transfer results between English and other languages. For tuned models the CKA also fails to reveal similarity for layers 6-11 (Figure \ref{fig:normformer_tuned} in Appendix \ref{sec:appendix}).

In this example, CKA is not capturing the notion of similarity that would coincide with zero-shot cross-lingual transfer performance for XLM-Normformer. Zero-shot transfer (say) from English requires language representations that \textit{converge} to English values so the other languages can re-use the linear prediction head (calibrated for English). % Note that we pre-train and identically fine-tune the models.

To double-check the result we also retrain the \verb|scale_normformer| the second time with a different random restart and get the same CKA results (see Figure \ref{fig:normformer_v2} in Appendix \ref{sec:appendix}). 

\begin{figure}
    \centering
    \includegraphics[width=0.38\textwidth]{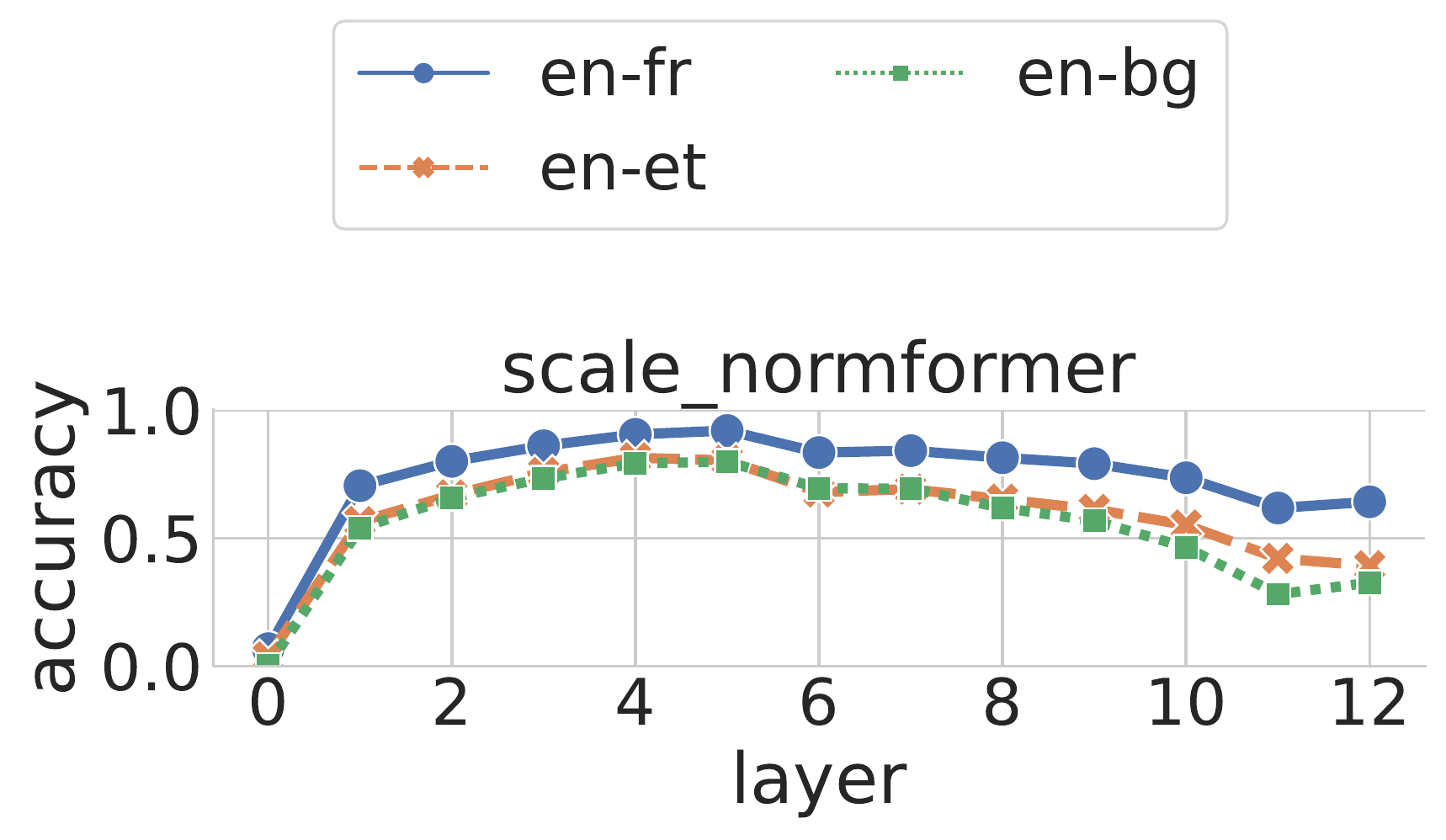}
    \caption{Per-layer sentence matching accuracy for the XLM-Normformer. The result again shows relatively high matching scores for the deeper layers in contrast to the CKA result from Figure \ref{fig:normformer_cka}. There is some decline, but nothing like ~zero similarity of CKA.}
    \label{fig:normformer_matching}
\end{figure}

\paragraph{Experiment 3: Per-Layer Matching Accuracy}
However, let us also see the results of our sentence matching task to verify whether these deep representations in Normformer are useful. Figure \ref{fig:normformer_matching} shows the resulting per-layer accuracy. 

The pattern shows that layers 6-12 show some significant cross-lingual matching scores (>50\% for French) with only a slightly decreasing trend. This experiment confirms that there are aspects of cross-lingual similarity in these multilingual representations that CKA failed to reveal.

\subsection{Downsides of CCA}
\label{sec:motiv-cca}

This section shows that the family of CCA-like similarity indexes suffers from similar issues as CKA.
The first downside is that CCA is hard to interpret. CCA is a second-order similarity index (similarly to CKA), which makes it hard to trace the reasons for high/low CCA scores to specific neurons or give any other fine-grained explanation. 
The second downside is that it is also not robust and has led to the misleading conclusion in the related literature (as demonstrated in \citealt{del-baltic}).
We discuss these downsides in more detail below.

\paragraph{Interpretability}
Another interesting aspect of our Normformer case is that PWCCA and SVCCA similarity indexes show correlations of about 0.5-0.8 for the layers 6-12 (see Figure \ref{fig:normformer_others} in Appendix \ref{sec:appendix} for verification). It indicates something special about CKA eigenvalue weighting, normalization (the denominator in Equation \ref{eq:cka}), or both. One possibility is that dominant eigenneurons (the ones that also have high eigenvalues)  in \textit{monolingual} representational spaces are unproportionally similar to each other (and this causes a high denominator and thus the low CKA scores). 

In any case, even if we recover what eigenvalues/normalization components cause these extremely low values, it would be even harder to track down which individual neurons cause the problem and to what extent (CCA/CKA methods essentially find linear combinations of the neurons and so mix them up). It highlights the interpretability issue with CKA/CCA indexes that arises  when these indexes disagree with our sanity check and with others. 

\paragraph{Conflicting Literature}
The disagreement between CCA/CKA also caused a problem of conflicting evidence in the literature. Namely, \citet{Singh2019BERTIN} used PWCCA to conclude that mBERT representations diverge starting from the early layers. However, this contradicts the evidence from the multiple behavior studies of mBERT that argue that the opposite is true \cite{wu-dredze-2019-beto, pires-etal-2019-multilingual, liu-etal-2020-multilingual, libovicky-etal-2020-language, conneau-etal-2020-emerging, muller-etal-2021-first}. \citet{del-baltic} find that merely changing the index from PWCCA to SVCCA or CKA in \cite{Singh2019BERTIN} produces results consistent with related works. It highlights the reliability issue with CKA/CCA. 

% Also, \citet{eq:cka} demonstrated superiority that CKA is superior to them (they also considered different layers of a single network which is somehow more similar to our setting).
In summary, similarity indexes value different aspects of representations and correspond to different concepts of similarity. It is, therefore, necessary to consult the specific analysis goal to define what we want the similarity to capture. It brings us to Section \ref{sec:method} where we propose a simple alternative method that aligns well with the goals of cross-lingual similarity analysis.

\section{Method: Average Neuron-Wise Correlation (ANC)}
\label{sec:method}
In Section \ref{sec:motiv} we demonstrated multiple drawbacks that CCA/CKA similarity indexes have in the cross-lingual context.

\subsection{Definition}
\paragraph{Assumption}
In this section, we propose a straightforward alternative method that builds on the assumption that neurons in representations for different languages are aligned one-to-one a priori. We find this assumption reasonable to make for several reasons. 

First, it aligns well with the goal that motivated most cross-lingual similarity analysis works: zero-shot cross-lingual transfer learning. Zero-shot transfer is possible because a linear prediction head fine-tuned (usually) for English can exploit \textbf{direct} linear relationships between English and (say) French representations. Indeed, the linear prediction head calibrates each weight to work with the specific English neuron. Having that specific neuron similar to the French neuron allows the linear head to work on French.

Second, it allows us to decompose the similarity index into correlations of individual neurons, thus facilitating interpretability. We can explicitly see which neurons contribute to the similarity the most/the least, and these neurons have an interpretation of being the most language-specific/language-natural. % /-natural ??

Third, it captures the most natural objectives that many cross-lingual alignment literature consider \cite{wu-dredze-2020-explicit}: representations of the same sentences should have the exact representations (in case the network is aligned). Residual connections strengthen this assumption for hidden layers.
    
\paragraph{Description}

The solution is straightforward: we compute individual correlations between pairs of English and (say French) neurons and calculate an average score. We also take absolute values of the correlations because the network can swap a negative correlation into a positive with a simple negative weight at the next layer.

Thus, we define Average Neuron-Wise Correlation (ANC) as follows.

Let the centered (by neurons) layer representations be
\begin{center}
    $X:=L_1 - mean(L_1)$\\
    $Y:=L_2 - mean(L_2)$
\end{center}

The  (Pearson) correlation $corr$ between two neurons $\vv{z_x}$ and $\vv{z_y}$ form $X$ and $Y$ is defined as:
\begin{align}
    corr(\vv{z_x}, \vv{z_y}) = \frac{\langle\vv{z_x}, \vv{z_y}\rangle}{\|\vv{z_x}\| \|\vv{z_y}\|}
\end{align}

We thus define The ANC similarity between two layers $L_1$ and $L_2$ as:

\begin{align}
    ANC(X,Y) = \frac{\sum_{i=1}^{n} abs(corr(\vv{z_x^i}, \vv{z_y^i}))}{n}
\end{align}

It is only possible for us to construct such an index because the neurons come from a single network where we already know what alignment between neurons is (and ought to be). The method will not work if neurons come from layers of two different networks, for example. In these cases, CCA-like indexes are likely the best fit.

\subsection{Sanity Checks}
\label{sec:method:sanity}
In this subsection, we verify that our method gives plausible predictions in the cases where we already know what the result should be.

% \paragraph{mBERT and XLM-R}
\paragraph{Based on the Insight From the Literature}
We based this sanity check on the known insight from the literature. The multilingual BERT model (\verb|bert-base-multilingual-cased|) is widely studied in the literature \cite{wu-dredze-2019-beto, pires-etal-2019-multilingual, liu-etal-2020-multilingual, conneau-etal-2020-emerging}. 
\citet{muller-etal-2021-first} provided direct behavioral evidence that representations in mBERT (\verb|bert-base-multilingual-cased|) should follow the ``first align, then predict'' pattern: they first converge towards each other and diverge slightly only at deep layers. 

\citet{libovicky-etal-2020-language} and \citet{del-baltic} demonstrated that the said pattern generalizes to the XLM-Roberta (\verb|xlm-roberta-base|) model \cite{xlmr}, which is similar in size and training objective to mBERT with the main differences being the removal of the next sentence prediction loss and training on the segments of texts (irrespectively to sentence boundaries)

So our method should reveal the ``first align, then predict'' pattern in these two cases. Otherwise, we conclude that it fails to capture the relevant properties of similarity we desire.

\begin{figure}[]
    \centering
    \includegraphics[width=0.34\textwidth]{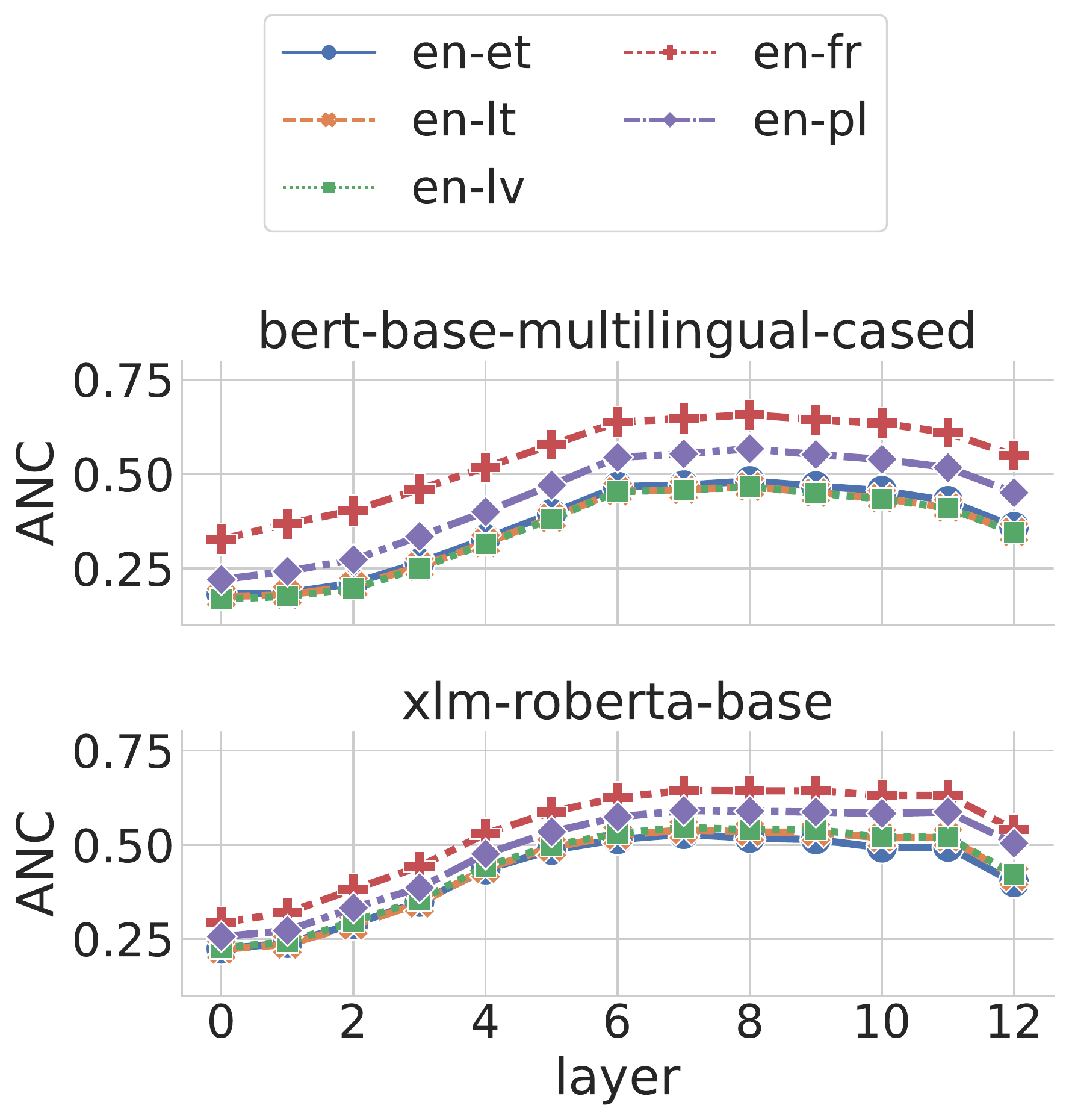}
    \caption{ANC result for the mBERT and XLM-R models. Our method captures the ``first align, then translate'' pattern presented in \citet{muller-etal-2021-first} and \citet{del-baltic}.}
    \label{fig:sanity_mbert_xlmr}
\end{figure}

Figure \ref{fig:sanity_mbert_xlmr} shows the resulting ANC scores for mBERT and XLM-R \verb|base| models.

The result demonstrates that our method passes the proposed sanity check by being able to reveal the ``first align, then predict'' pattern. Also, the correlation at the most language natural layers is about 0.7, which indicates that the ANC's \textit{strong assumption} of one-to-one aligned neurons is informative. Lastly, we can see that the ANC distance between English and other languages is more considerable for mBERT than for XLM-R, which corresponds to how these models perform in a cross-lingual transfer \cite{xlmr}.

% \paragraph{XLM-Normformer}
\paragraph{Based on the Experiment in Section \ref{sec:motiv}}
We base this sanity check on the same XLM-Roberta Normformer experiment that we used to present the CKA failure case in Section \ref{sec:motiv}. Our method should be able to reveal that representations at deeper layers in \verb|scale_normformer| are somehow cross-lingually similar. Moreover, it should also keep the results for the analogous \verb|scale_post| and \verb|scale_pre models| models in agreement.    

\begin{figure}[]
    \centering
    \includegraphics[width=0.34\textwidth]{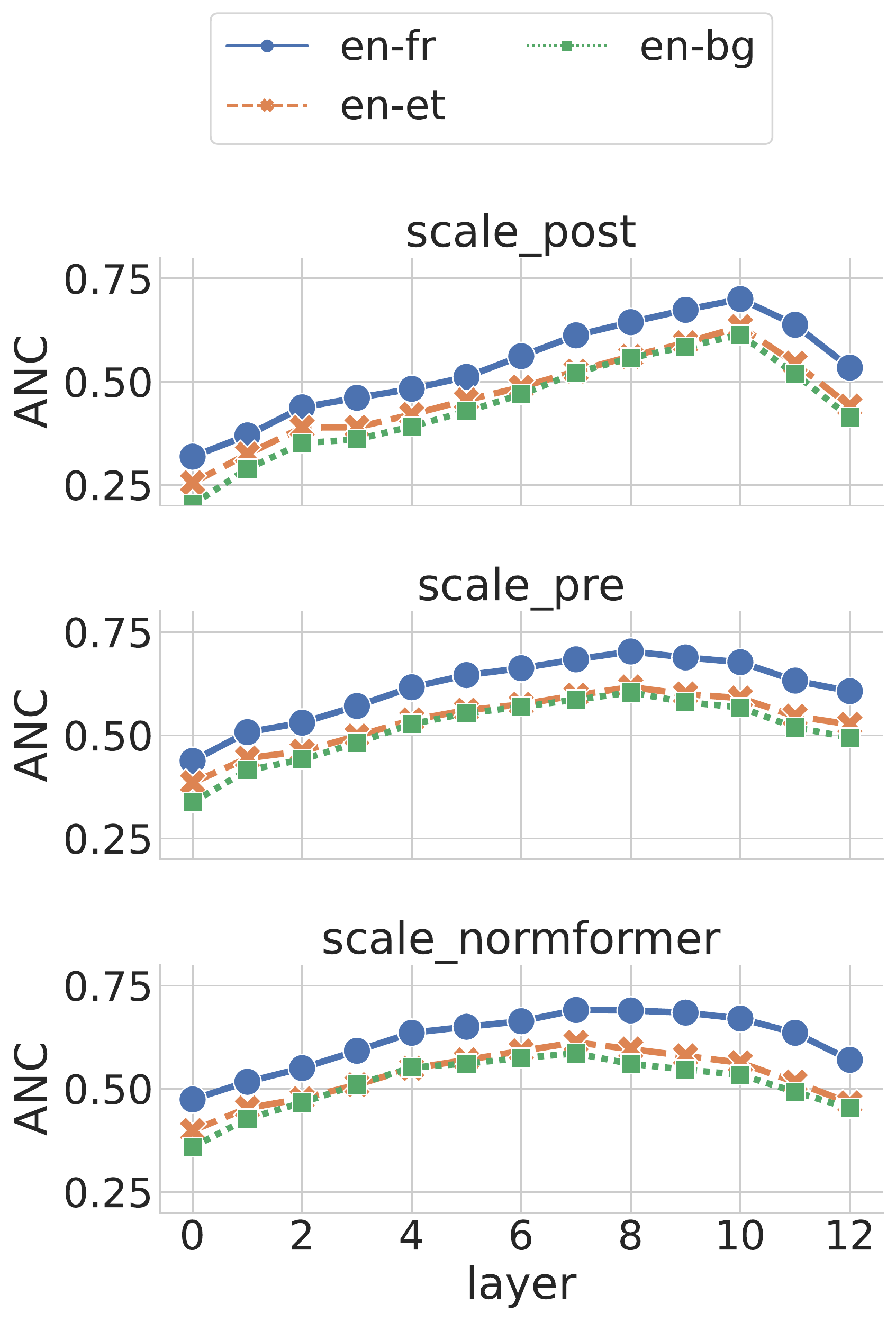}
    \caption{ANC result for the three models we presented in Section \ref{sec:motiv}. Our method, unlike CKA (Figure~\ref{fig:normformer_cka}), does capture the cross-lingual similarity existing in the deeper layers of XLM-Roberta Normformer (\textit{scale\_normformer}).}
    \label{fig:normfromer_anc}
\end{figure}

We present ANC results for the Section \ref{sec:motiv} experiment in Figure \ref{fig:normfromer_anc}.

The figure shows that unlike CKA (Figure \ref{fig:normformer_cka}), the ANC is able to reveal the ``first align, then predict'' pattern for the \verb|scale_normformer| and better explains the evidence we provided in Table \ref{tab:xnli} and Figure \ref{fig:normformer_matching}.   

In summary, this section demonstrated that our method passes the sanity checks of both related literature and the Section \ref{sec:motiv} experiment (that made CKA fail). In addition, considering how simple it is to interpret ANC scores (the score is a simple average of neuron-wise correlations), the method is a beneficial tool for comparing representation between languages in a single multilingual model.

\section{Scaling Laws of Cross-lingual Representational Similarity in Multilingual Models}
\label{sec:scaling_laws}
In previous sections, we justified our claim that ANC is better suited for cross-lingual analysis than CCA/CKA methods. In this section, we present an application of ANC to the analysis of representational similarity scaling in cross-lingual language models.

Most related works that analyzed representational patterns in multilingual language models focused on a single model, such as \verb|base| version of mBERT or XLM-R. In Section \ref{sec:method:sanity} we covered these models showing that ANC accompanies our representational similarity index demands from these models. However, as the model scaling brings significant improvements in downstream tasks performance, we must focus our analysis efforts on the large models and scaling laws \cite{bowman-2022-dangers}.
 In this section, we use ANC to explore if the ``first align, then predict'' pattern generalizes to CLMs and if it preserves in the large-scale versions of multilingual MLMs and CLMs.

\begin{table}[]
\centering
\small
\begin{tabular}{@{}lccccc@{}}
\toprule
\multicolumn{1}{c}{Name} & type & \#params & l  & n    & \#lgs \\ \midrule
xlm-roberta-base         & MLM & 270M    & 12 & 758  & 100  \\
xlm-roberta-large        & MLM & 550M    & 24 & 1024 & 100  \\
xlm-roberta-xl           & MLM & 3.5B    & 36 & 2560 & 100  \\
xlm-roberta-xxl          & MLM & 10.7B   & 48 & 4096 & 100  \\
\toprule
xglm-564M & CLM & 564M   & 24     & 1024      & 30    \\
xglm-1.7B & CLM & 1.7B   & 24     & 2048      & 30    \\
xglm-2.9B & CLM & 2.9B   & 48     & 2048      & 30    \\
xglm-4.5B & CLM & 4.5B   & 48     & 4096      & 134   \\
xglm-7.5B & CLM & 7.5B   & 32     & 4096      & 30    \\ \bottomrule
\end{tabular}
\caption{Model details for XLM-R and XGLM models we study. \emph{type}: training objective of the model, \emph{\#params}: number of parameters, $l$: number of layers, $n$: number of hidden units (neurons at each layer), \emph{\#lgs}: number of languages used in pertaining.}
\label{tab:scaling_laws}
\end{table}

\paragraph{Model Details}
We describe the models we study in Table \ref{tab:scaling_laws}. The Table shows that there are two groups of models: MLMs (encoder only) and CLMs (decoder only). Models in each group notably vary in a number of parameters and neurons at each layer.

\begin{figure}[H]
    \centering
    \includegraphics[width=0.43\textwidth]{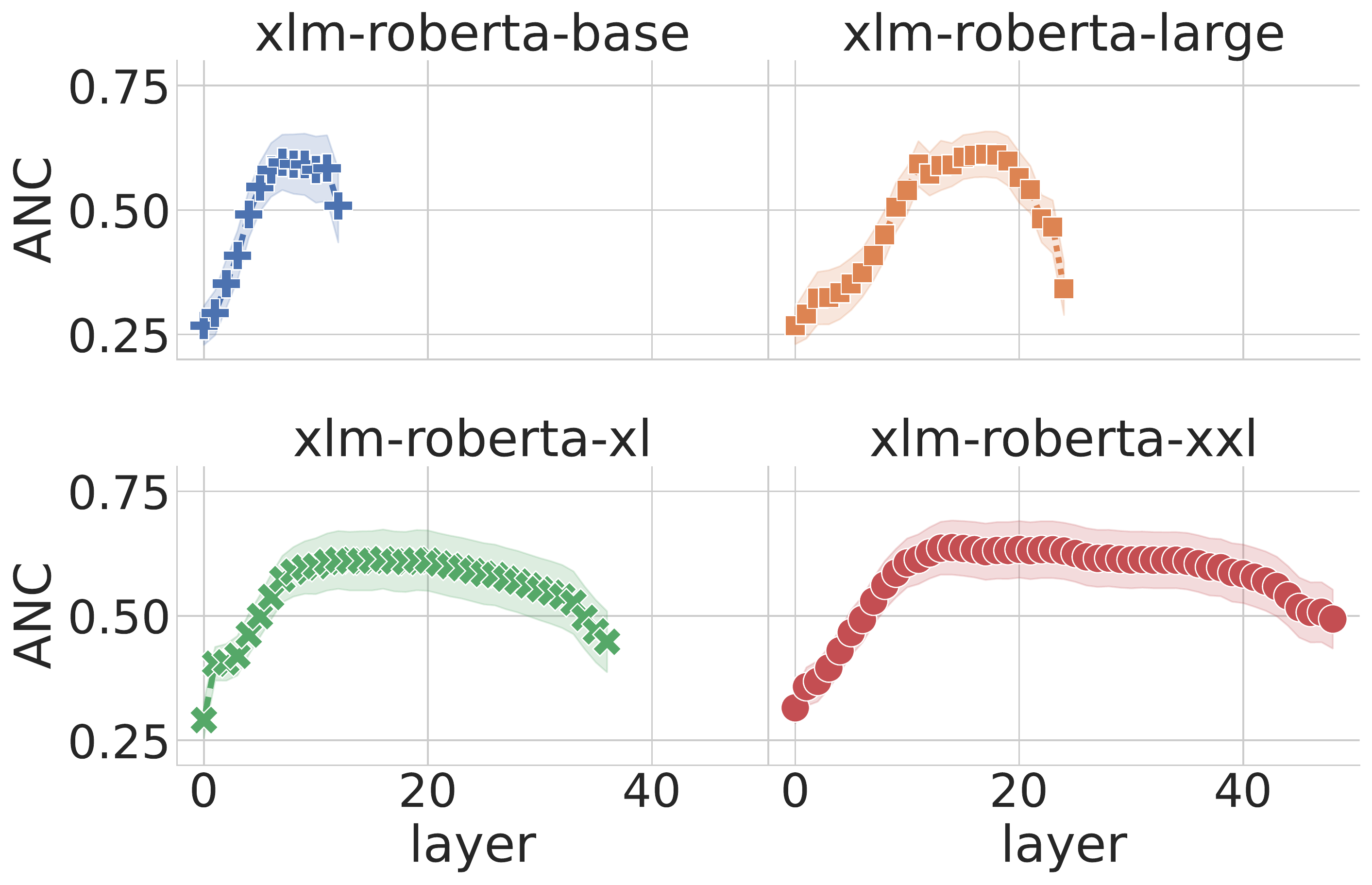}
    \caption{ANC cross-lingual representational similarity for the XLM-R MLM-style models of different sizes. All models follow a similar ``first align, then predict'' pattern. We aggregate among en-fr, en-de, en-ru, and en-et pairs and show similarity average and spread.}
    \label{fig:laws-xlmr}
\end{figure}

\begin{figure}[H]
    \centering
    \includegraphics[width=0.43\textwidth]{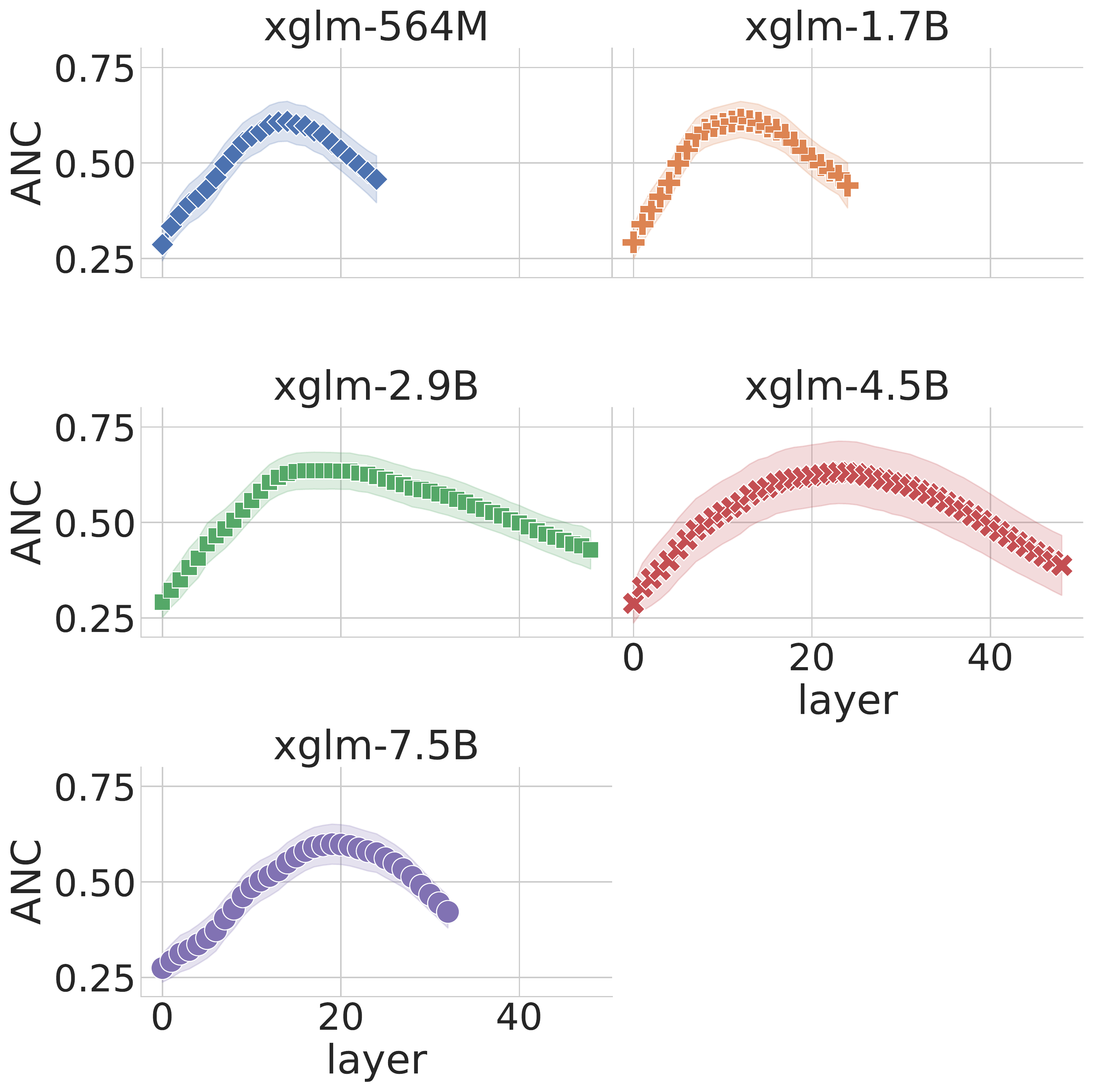}
    \caption{ANC cross-lingual representational similarity for the XGLM CLM-style models of different sizes. All models follow a similar ``first align, then predict'' pattern. We aggregate among en-fr, en-de, en-ru, and en-et pairs and show similarity average and spread.}
    \label{fig:laws-xglm}
\end{figure}

\paragraph{Results}
Figures \ref{fig:laws-xlmr} and \ref{fig:laws-xglm} reveal that the cross-lingual similarity of multilingual representations in all the networks we study follows the same ``first align, then translate'' pattern. It happens despite differences in training objectives, number of languages, and sizes. Therefore, this result provides evidence that multilingual models rely on the exact mechanism described in \cite{muller-etal-2021-first}, independently of the size or the MLM/CLM training objective.

\section{Conclusion}
In this study, we introduced an example where \textit{CKA} drastically fails to reveal the cross-lingual similarity between language representations across the deeper layers of the multilingual model. We also highlighted that \textit{CCA} methods suffer from related problems as well (despite passing that concrete sanity check that CKA failed). 

Then, we proposed a new approach: Average Neuron-Wise Correlation (ANC), which builds on the assumption of neuron alignment in cross-lingual representations. We verified that our method passes the sanity check at which CKA fails and produces results harmonious with the evidence from related work. 

Finally, we used ANC to show that the ``first align, then translate'' pattern of cross-lingual representations generalizes to CLMs and the larger scales of MLMs and CLMs.

\section*{Acknowledgements}
This work has been supported by the grant No. 825303 (Bergamot\footnote{\url{https://browser.mt/}}) of European Union’s Horizon 2020 research and innovation program.

We also thank the Unversity of Tartu's High-Performance Computing Center 
for providing GPU computing resources \cite{tartuhpc}.

\section*{Ethical Considerations}
Our work aims to improve the methodology used to perform cross-lingual similarity analysis in multilingual models. We showed that our method outperforms the previous tooling across languages sampled from the four language families: Germanic, Romance, Slavic, Baltic, and Finno-Ugric. To introduce more diversity, we sample different languages from these language families in different experiments. However, we did not experiment with other language families and extremely low-resource languages. These languages might be underrepresented in pretrained LMs and require different analysis tooling. At the time being, we recommend using our method \textit{together} with the previous methods for more reliable results in these cases.

\bibliography{anthology,custom}

\begin{thebibliography}{25}
\expandafter\ifx\csname natexlab\endcsname\relax\def\natexlab#1{#1}\fi

\bibitem[{Bowman(2022)}]{bowman-2022-dangers}
Samuel Bowman. 2022.
\newblock \href {https://doi.org/10.18653/v1/2022.acl-long.516} {The dangers of
  underclaiming: Reasons for caution when reporting how {NLP} systems fail}.
\newblock In \emph{Proceedings of the 60th Annual Meeting of the Association
  for Computational Linguistics (Volume 1: Long Papers)}, pages 7484--7499,
  Dublin, Ireland. Association for Computational Linguistics.

\bibitem[{Brown et~al.(2020)Brown, Mann, Ryder, Subbiah, Kaplan, Dhariwal,
  Neelakantan, Shyam, Sastry, Askell, Agarwal, Herbert-Voss, Krueger, Henighan,
  Child, Ramesh, Ziegler, Wu, Winter, Hesse, Chen, Sigler, Litwin, Gray, Chess,
  Clark, Berner, McCandlish, Radford, Sutskever, and
  Amodei}]{gpt-NEURIPS2020_1457c0d6}
Tom Brown, Benjamin Mann, Nick Ryder, Melanie Subbiah, Jared~D Kaplan, Prafulla
  Dhariwal, Arvind Neelakantan, Pranav Shyam, Girish Sastry, Amanda Askell,
  Sandhini Agarwal, Ariel Herbert-Voss, Gretchen Krueger, Tom Henighan, Rewon
  Child, Aditya Ramesh, Daniel Ziegler, Jeffrey Wu, Clemens Winter, Chris
  Hesse, Mark Chen, Eric Sigler, Mateusz Litwin, Scott Gray, Benjamin Chess,
  Jack Clark, Christopher Berner, Sam McCandlish, Alec Radford, Ilya Sutskever,
  and Dario Amodei. 2020.
\newblock \href
  {https://proceedings.neurips.cc/paper/2020/file/1457c0d6bfcb4967418bfb8ac142f64a-Paper.pdf}
  {Language models are few-shot learners}.
\newblock In \emph{Advances in Neural Information Processing Systems},
  volume~33, pages 1877--1901. Curran Associates, Inc.

\bibitem[{Conneau and Lample(2019)}]{xlmr}
Alexis Conneau and Guillaume Lample. 2019.
\newblock \emph{Cross-Lingual Language Model Pretraining}, chapter~33. Curran
  Associates Inc., Red Hook, NY, USA.

\bibitem[{Conneau et~al.(2018)Conneau, Rinott, Lample, Williams, Bowman,
  Schwenk, and Stoyanov}]{conneau2018xnli}
Alexis Conneau, Ruty Rinott, Guillaume Lample, Adina Williams, Samuel~R.
  Bowman, Holger Schwenk, and Veselin Stoyanov. 2018.
\newblock Xnli: Evaluating cross-lingual sentence representations.
\newblock In \emph{Proceedings of the 2018 Conference on Empirical Methods in
  Natural Language Processing}. Association for Computational Linguistics.

\bibitem[{Conneau et~al.(2020)Conneau, Wu, Li, Zettlemoyer, and
  Stoyanov}]{conneau-etal-2020-emerging}
Alexis Conneau, Shijie Wu, Haoran Li, Luke Zettlemoyer, and Veselin Stoyanov.
  2020.
\newblock \href {https://doi.org/10.18653/v1/2020.acl-main.536} {Emerging
  cross-lingual structure in pretrained language models}.
\newblock In \emph{Proceedings of the 58th Annual Meeting of the Association
  for Computational Linguistics}, pages 6022--6034, Online. Association for
  Computational Linguistics.

\bibitem[{Del and Fishel(2021)}]{del-baltic}
Maksym Del and Mark Fishel. 2021.
\newblock \href {https://doi.org/10.48550/ARXIV.2109.01207} {{Similarity of
  Sentence Representations in Multilingual LMs: Resolving Conflicting
  Literature and Case Study of Baltic Languages}}.
\newblock arXiv.

\bibitem[{Hotelling(1936)}]{cca-Ramsay1984MatrixC}
Harold Hotelling. 1936.
\newblock \href {https://doi.org/10.1093/biomet/28.3-4.321} {{Relations Between
  Two Sets Of Variates*}}.
\newblock \emph{Biometrika}, 28(3-4):321--377.

\bibitem[{Kornblith et~al.(2019)Kornblith, Norouzi, Lee, and
  Hinton}]{cka=pmlr-v97-kornblith19a}
Simon Kornblith, Mohammad Norouzi, Honglak Lee, and Geoffrey Hinton. 2019.
\newblock \href {https://proceedings.mlr.press/v97/kornblith19a.html}
  {Similarity of neural network representations revisited}.
\newblock In \emph{Proceedings of the 36th International Conference on Machine
  Learning}, volume~97 of \emph{Proceedings of Machine Learning Research},
  pages 3519--3529. PMLR.

\bibitem[{Kudugunta et~al.(2019)Kudugunta, Bapna, Caswell, and
  Firat}]{mt-svcca}
Sneha Kudugunta, Ankur Bapna, Isaac Caswell, and Orhan Firat. 2019.
\newblock \href {https://doi.org/10.18653/v1/d19-1167} {Investigating
  multilingual nmt representations at scale}.
\newblock \emph{Proceedings of the 2019 Conference on Empirical Methods in
  Natural Language Processing and the 9th International Joint Conference on
  Natural Language Processing (EMNLP-IJCNLP)}.

\bibitem[{Li et~al.(2015)Li, Yosinski, Clune, Lipson, and
  Hopcroft}]{Li2015ConvergentLD}
Yixuan Li, Jason Yosinski, Jeff Clune, Hod Lipson, and John~E. Hopcroft. 2015.
\newblock Convergent learning: Do different neural networks learn the same
  representations?
\newblock In \emph{FE@NIPS}.

\bibitem[{Libovick{\'y} et~al.(2020)Libovick{\'y}, Rosa, and
  Fraser}]{libovicky-etal-2020-language}
Jind{\v{r}}ich Libovick{\'y}, Rudolf Rosa, and Alexander Fraser. 2020.
\newblock \href {https://doi.org/10.18653/v1/2020.findings-emnlp.150} {On the
  language neutrality of pre-trained multilingual representations}.
\newblock In \emph{Findings of the Association for Computational Linguistics:
  EMNLP 2020}, pages 1663--1674, Online. Association for Computational
  Linguistics.

\bibitem[{Lin et~al.(2021)Lin, Mihaylov, Artetxe, Wang, Chen, Simig, Ott,
  Goyal, Bhosale, Du, Pasunuru, Shleifer, Koura, Chaudhary, O'Horo, Wang,
  Zettlemoyer, Kozareva, Diab, Stoyanov, and Li}]{xglm}
Xi~Victoria Lin, Todor Mihaylov, Mikel Artetxe, Tianlu Wang, Shuohui Chen,
  Daniel Simig, Myle Ott, Naman Goyal, Shruti Bhosale, Jingfei Du, Ramakanth
  Pasunuru, Sam Shleifer, Punit~Singh Koura, Vishrav Chaudhary, Brian O'Horo,
  Jeff Wang, Luke Zettlemoyer, Zornitsa Kozareva, Mona Diab, Veselin Stoyanov,
  and Xian Li. 2021.
\newblock \href {https://doi.org/10.48550/ARXIV.2112.10668} {Few-shot learning
  with multilingual language models}.
\newblock arXiv.

\bibitem[{Liu et~al.(2020)Liu, Zhang, Zhang, Singh, Saraf, and
  Zweig}]{liu-etal-2020-multilingual}
Chunxi Liu, Qiaochu Zhang, Xiaohui Zhang, Kritika Singh, Yatharth Saraf, and
  Geoffrey Zweig. 2020.
\newblock \href {https://aclanthology.org/2020.sltu-1.7} {Multilingual
  graphemic hybrid {ASR} with massive data augmentation}.
\newblock In \emph{Proceedings of the 1st Joint Workshop on Spoken Language
  Technologies for Under-resourced languages (SLTU) and Collaboration and
  Computing for Under-Resourced Languages (CCURL)}, pages 46--52, Marseille,
  France. European Language Resources association.

\bibitem[{Morcos et~al.(2018)Morcos, Raghu, and Bengio}]{pwcca-NIPS2018_7815}
Ari Morcos, Maithra Raghu, and Samy Bengio. 2018.
\newblock \href
  {http://papers.nips.cc/paper/7815-insights-on-representational-similarity-in-neural-networks-with-canonical-correlation.pdf}
  {Insights on representational similarity in neural networks with canonical
  correlation}.
\newblock In S.~Bengio, H.~Wallach, H.~Larochelle, K.~Grauman, N.~Cesa-Bianchi,
  and R.~Garnett, editors, \emph{Advances in Neural Information Processing
  Systems 31}, pages 5732--5741. Curran Associates, Inc.

\bibitem[{Muller et~al.(2021)Muller, Elazar, Sagot, and
  Seddah}]{muller-etal-2021-first}
Benjamin Muller, Yanai Elazar, Beno{\^\i}t Sagot, and Djam{\'e} Seddah. 2021.
\newblock \href {https://doi.org/10.18653/v1/2021.eacl-main.189} {First align,
  then predict: Understanding the cross-lingual ability of multilingual
  {BERT}}.
\newblock In \emph{Proceedings of the 16th Conference of the European Chapter
  of the Association for Computational Linguistics: Main Volume}, pages
  2214--2231, Online. Association for Computational Linguistics.

\bibitem[{Pires et~al.(2019)Pires, Schlinger, and
  Garrette}]{pires-etal-2019-multilingual}
Telmo Pires, Eva Schlinger, and Dan Garrette. 2019.
\newblock \href {https://doi.org/10.18653/v1/P19-1493} {How multilingual is
  multilingual {BERT}?}
\newblock In \emph{Proceedings of the 57th Annual Meeting of the Association
  for Computational Linguistics}, pages 4996--5001, Florence, Italy.
  Association for Computational Linguistics.

\bibitem[{Raghu et~al.(2017)Raghu, Gilmer, Yosinski, and
  Sohl-Dickstein}]{svcca-NIPS2017_7188}
Maithra Raghu, Justin Gilmer, Jason Yosinski, and Jascha Sohl-Dickstein. 2017.
\newblock \href
  {http://papers.nips.cc/paper/7188-svcca-singular-vector-canonical-correlation-analysis-for-deep-learning-dynamics-and-interpretability.pdf}
  {Svcca: Singular vector canonical correlation analysis for deep learning
  dynamics and interpretability}.
\newblock In I.~Guyon, U.~V. Luxburg, S.~Bengio, H.~Wallach, R.~Fergus,
  S.~Vishwanathan, and R.~Garnett, editors, \emph{Advances in Neural
  Information Processing Systems 30}, pages 6076--6085. Curran Associates, Inc.

\bibitem[{Shleifer et~al.(2021)Shleifer, Weston, and
  Ott}]{normformer-https://doi.org/10.48550/arxiv.2110.09456}
Sam Shleifer, Jason Weston, and Myle Ott. 2021.
\newblock \href {https://doi.org/10.48550/ARXIV.2110.09456} {Normformer:
  Improved transformer pretraining with extra normalization}.
\newblock arXiv.

\bibitem[{Singh et~al.(2019{\natexlab{a}})Singh, McCann, Socher, and
  Xiong}]{Singh2019BERTIN}
Jasdeep Singh, Bryan McCann, Richard Socher, and Caiming Xiong.
  2019{\natexlab{a}}.
\newblock Bert is not an interlingua and the bias of tokenization.
\newblock In \emph{EMNLP}.

\bibitem[{Singh et~al.(2019{\natexlab{b}})Singh, McCann, Socher, and
  Xiong}]{singh-etal-2019-bert}
Jasdeep Singh, Bryan McCann, Richard Socher, and Caiming Xiong.
  2019{\natexlab{b}}.
\newblock \href {https://doi.org/10.18653/v1/D19-6106} {{BERT} is not an
  interlingua and the bias of tokenization}.
\newblock In \emph{Proceedings of the 2nd Workshop on Deep Learning Approaches
  for Low-Resource NLP (DeepLo 2019)}, pages 47--55, Hong Kong, China.
  Association for Computational Linguistics.

\bibitem[{{University of Tartu}(2018)}]{tartuhpc}
{University of Tartu}. 2018.
\newblock \href {https://doi.org/10.23673/PH6N-0144} {Ut rocket}.

\bibitem[{Vaswani et~al.(2017)Vaswani, Shazeer, Parmar, Uszkoreit, Jones,
  Gomez, Kaiser, and Polosukhin}]{trans-NIPS2017_3f5ee243}
Ashish Vaswani, Noam Shazeer, Niki Parmar, Jakob Uszkoreit, Llion Jones,
  Aidan~N Gomez, \L~ukasz Kaiser, and Illia Polosukhin. 2017.
\newblock \href
  {https://proceedings.neurips.cc/paper/2017/file/3f5ee243547dee91fbd053c1c4a845aa-Paper.pdf}
  {Attention is all you need}.
\newblock In \emph{Advances in Neural Information Processing Systems},
  volume~30. Curran Associates, Inc.

\bibitem[{Wu and Dredze(2019)}]{wu-dredze-2019-beto}
Shijie Wu and Mark Dredze. 2019.
\newblock \href {https://doi.org/10.18653/v1/D19-1077} {Beto, bentz, becas: The
  surprising cross-lingual effectiveness of {BERT}}.
\newblock In \emph{Proceedings of the 2019 Conference on Empirical Methods in
  Natural Language Processing and the 9th International Joint Conference on
  Natural Language Processing (EMNLP-IJCNLP)}, pages 833--844, Hong Kong,
  China. Association for Computational Linguistics.

\bibitem[{Wu and Dredze(2020)}]{wu-dredze-2020-explicit}
Shijie Wu and Mark Dredze. 2020.
\newblock \href {https://doi.org/10.18653/v1/2020.emnlp-main.362} {Do explicit
  alignments robustly improve multilingual encoders?}
\newblock In \emph{Proceedings of the 2020 Conference on Empirical Methods in
  Natural Language Processing (EMNLP)}, pages 4471--4482, Online. Association
  for Computational Linguistics.

\bibitem[{Xiong et~al.(2020)Xiong, Yang, He, Zheng, Zheng, Xing, Zhang, Lan,
  Wang, and Liu}]{preln-10.5555/3524938.3525913}
Ruibin Xiong, Yunchang Yang, Di~He, Kai Zheng, Shuxin Zheng, Chen Xing,
  Huishuai Zhang, Yanyan Lan, Liwei Wang, and Tie-Yan Liu. 2020.
\newblock On layer normalization in the transformer architecture.
\newblock In \emph{Proceedings of the 37th International Conference on Machine
  Learning}, ICML'20. JMLR.org.

\end{thebibliography}
\bibliographystyle{acl_natbib}
\appendix
\section{Appendix}
\label{sec:appendix}
This appendix contains supplementary figures that support some auxiliary claims throughout the paper.

\begin{figure}[H]
    \centering
    \includegraphics[width=0.4\textwidth]{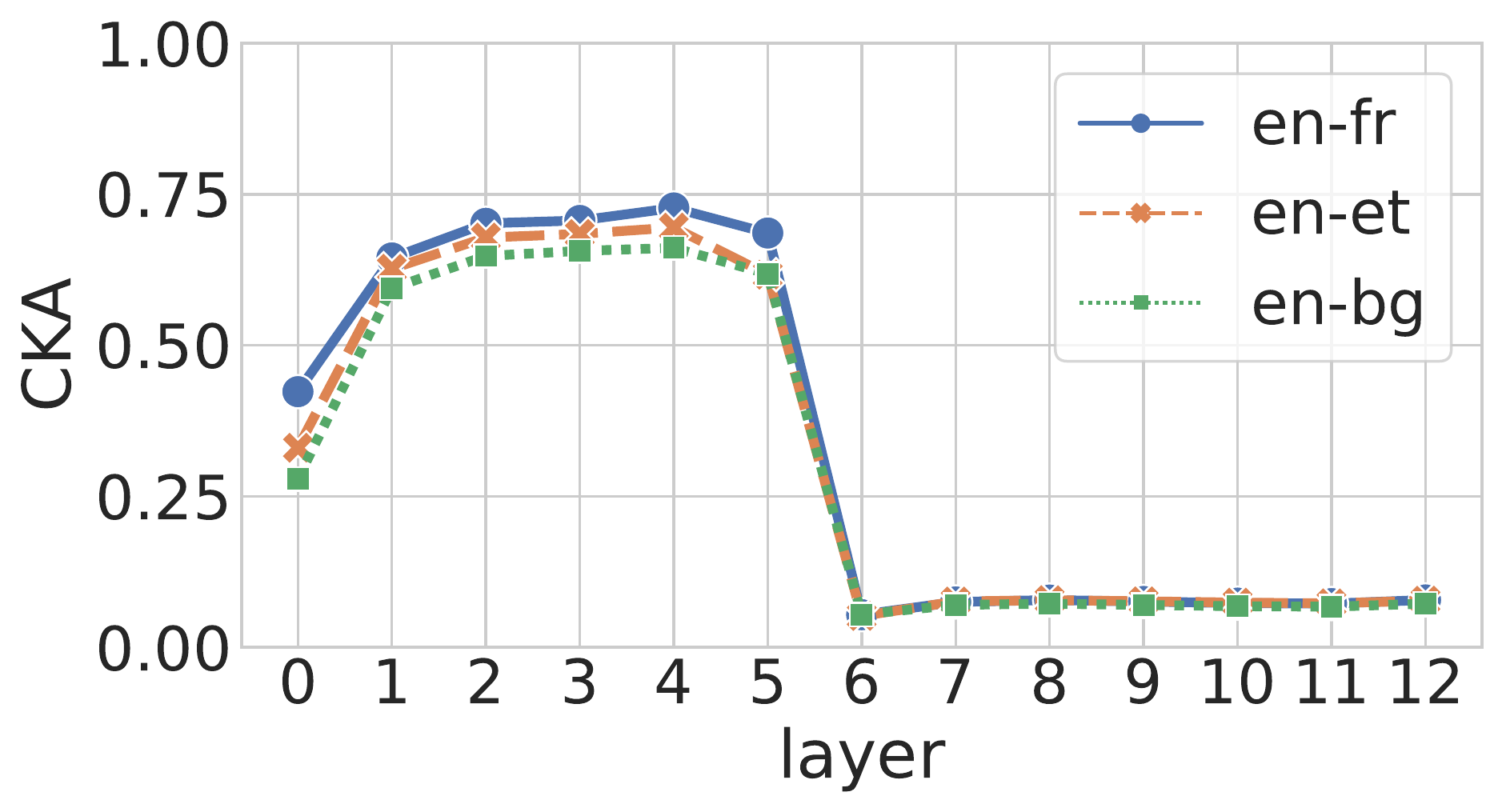}
    \caption{The CKA score for another Normformer (\textit{scale normformer}) model that we pre-trained from the different initialization. The cross-lingual similarity of deeper layers is about zero according to CKA despite evidence of the opposite from Section \ref{sec:motiv:results}}
    \label{fig:normformer_v2} 
\end{figure}

\begin{figure}[H]
    \centering
    \includegraphics[width=0.4\textwidth]{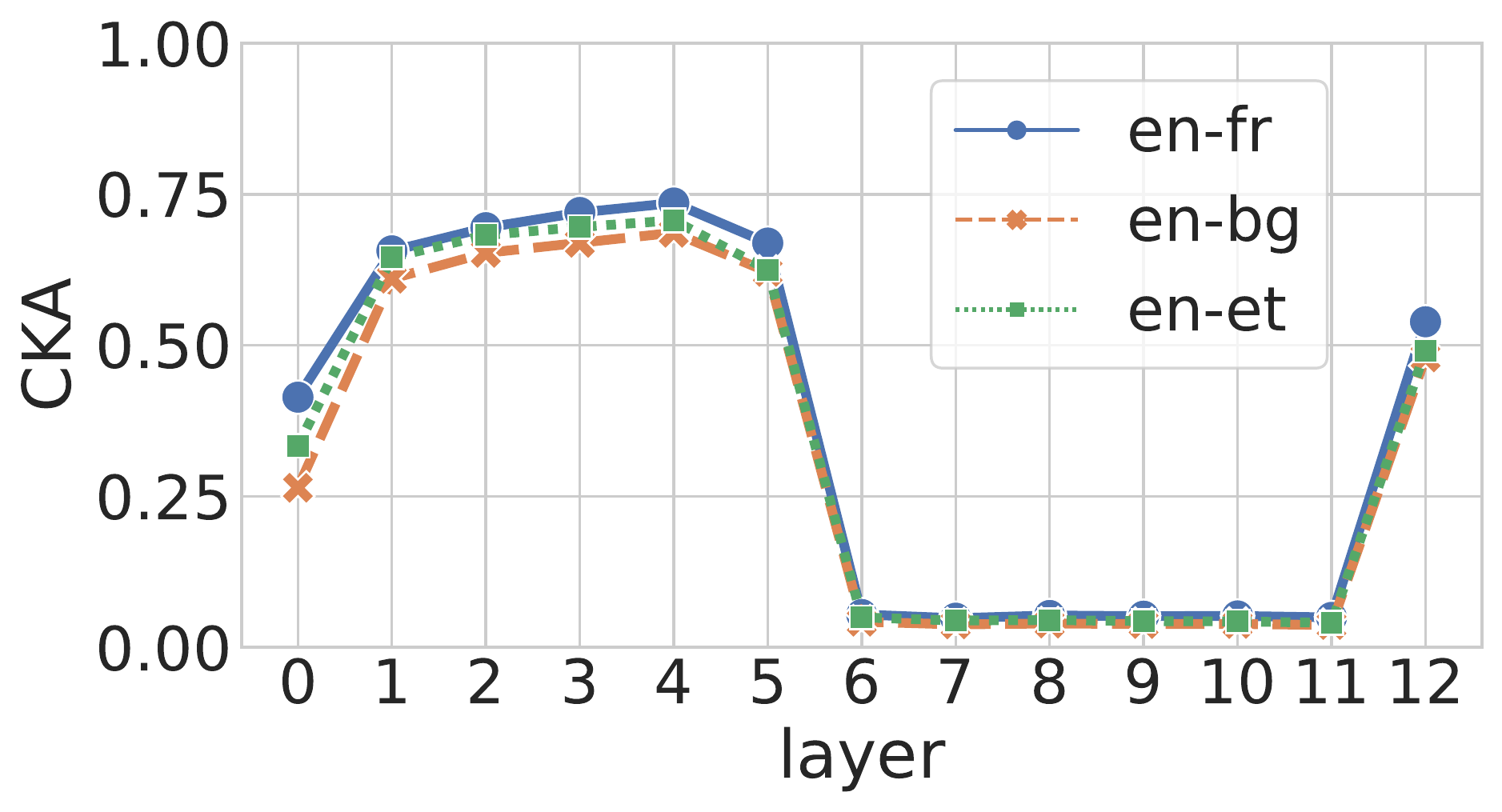}
    \includegraphics[width=0.4\textwidth]{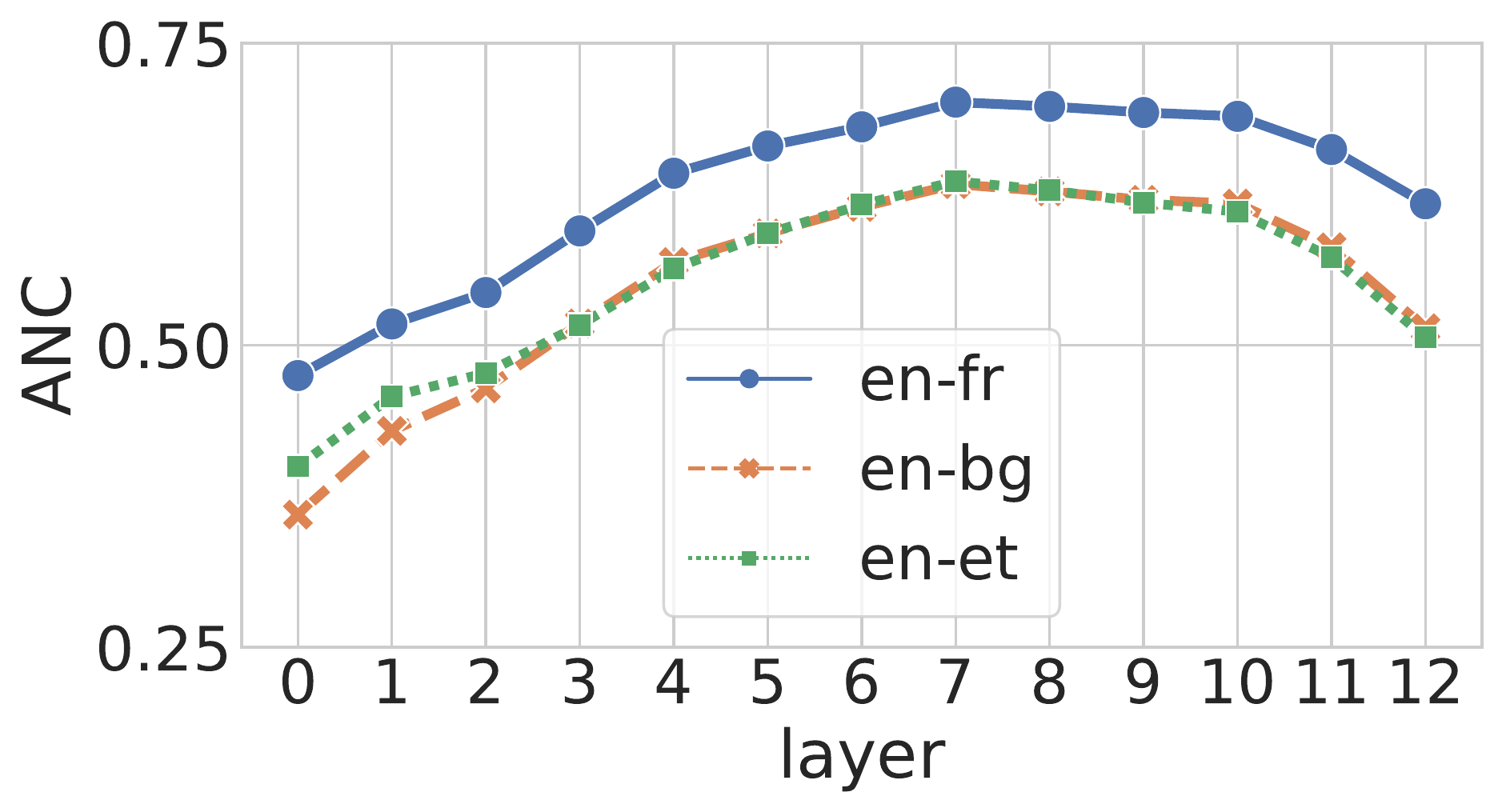}

    \caption{CKA and ANC results for the XLM-Normformer tuned on XNLI. The last layer is a CLS-pooled embedding (the one we tune for XNLI), while others are mean-poolings. CKA captures the similarity between CLS representations at the last layer but fails to capture it at layers 6-11. ANC captures the similarity across all layers.}
    \label{fig:normformer_tuned} 
\end{figure}

\begin{figure}[H]
    \centering
    \includegraphics[width=0.4\textwidth]{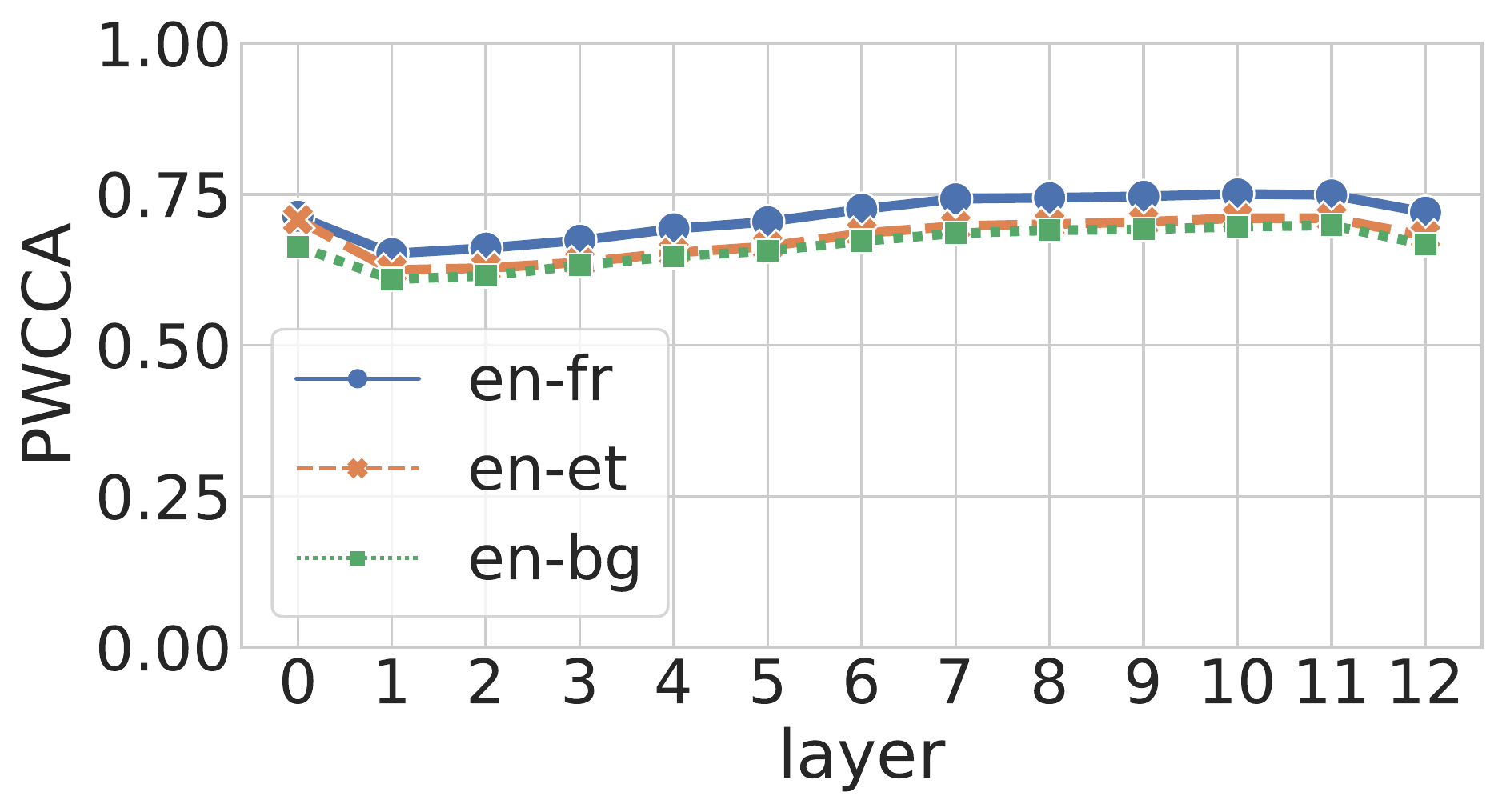}
    \includegraphics[width=0.45\textwidth]{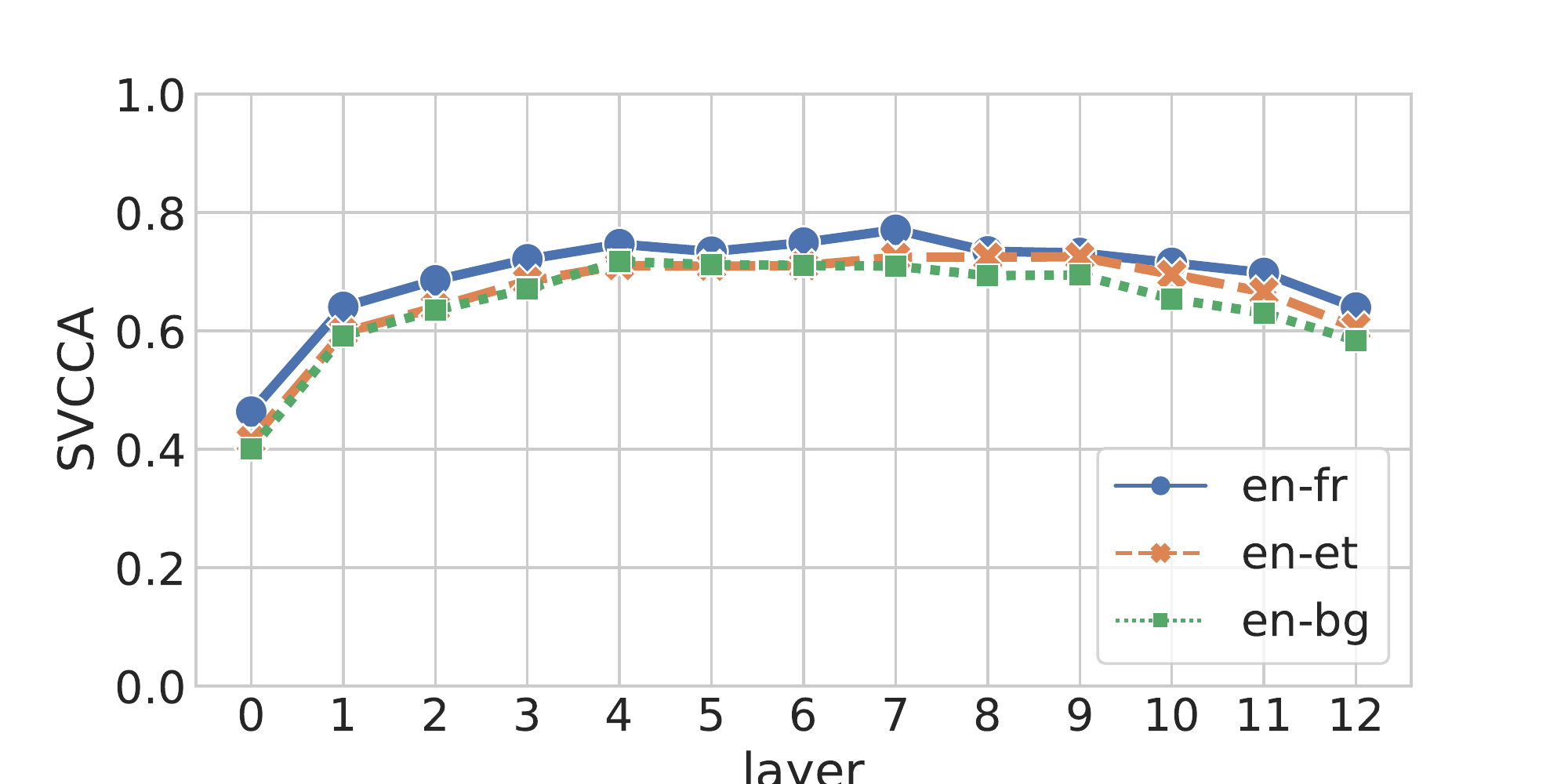}

    \caption{PWCCA and SVCCA results for the XLM-Normformer. These results are more intuitive to our notion of similarity for this particular case but struggle in other scenarios. }
    \label{fig:normformer_others} 
\end{figure}

\end{document}